\documentclass[10pt,twocolumn,letterpaper]{article}

\usepackage[pagenumbers]{cvpr}
\usepackage{amssymb}

\usepackage{url}
\usepackage{threeparttable}
\usepackage{makecell}

\definecolor{cvprblue}{rgb}{0.21,0.49,0.74}
\usepackage[pagebackref,breaklinks,colorlinks,allcolors=cvprblue]{hyperref}

\def\confName{CVPR}
\def\confYear{2027}

\title{Reality Perception Gap: Are AI-Manipulated Video Detectors Ready for the Real World?
}

\author{%
\parbox{0.95\textwidth}{\centering
Xiangrui Liu$^{1,*}$  \quad
Xinmiao Luan$^{1,*}$ \quad
Xinyuan Wang$^{1}$\\
Zhiyuan Fang$^{1}$ \quad
Chi-Yao Huang$^{1}$ \quad
Douglas C. Montgomery$^{1}$\\
Yanjie Fu$^{1}$ \quad
Felix Juefei-Xu$^{2}$ \quad
Yezhou Yang$^{1}$\\[6pt]
{\small
$^{1}$Arizona State University \quad
$^{2}$New York University
}\\[4pt]
{\ttfamily\small xiangrui@asu.edu}
}}

\begin{document}

\twocolumn[{%
\renewcommand\twocolumn[1][]{#1}%
\maketitle

\begin{center}
    \centering
    \captionsetup{type=figure}
    \includegraphics[width=0.9\textwidth]{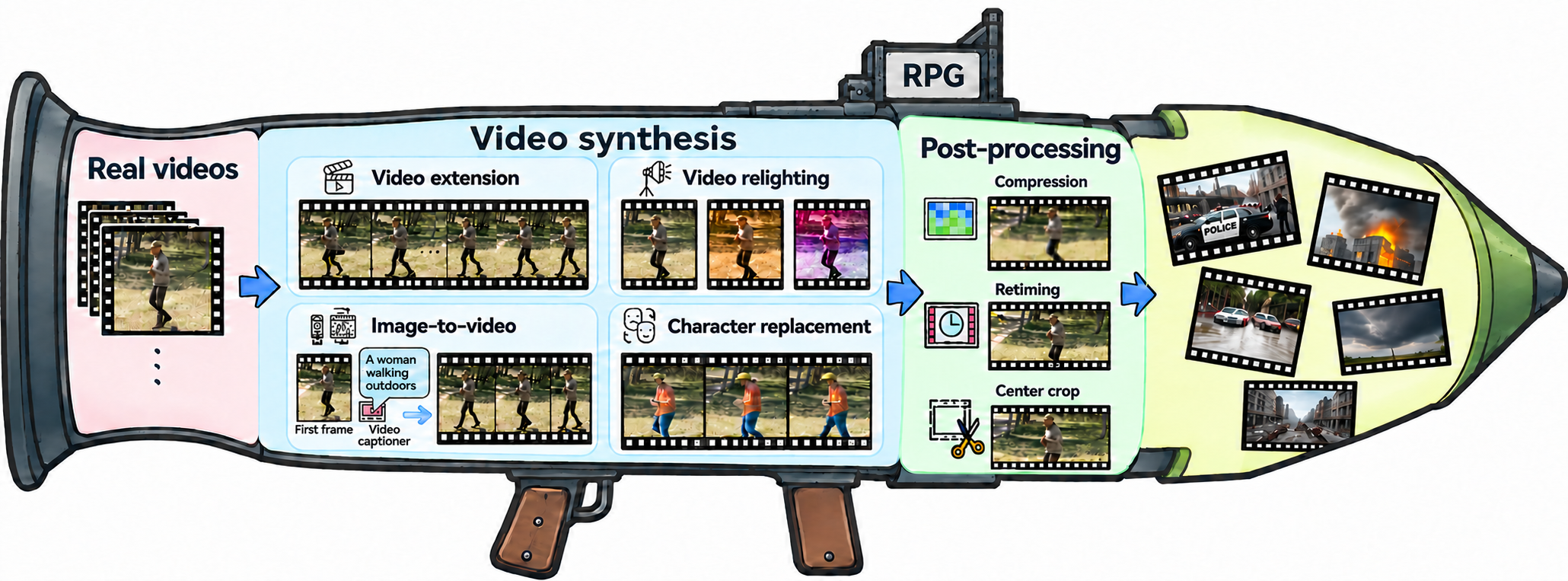}
    \captionof{figure}{Overview of the RPG benchmark construction pipeline. Real videos are routed through one of four synthesis operations, then passed through a post processing stage that mimics the transformations videos undergo when redistributed online and shields the benchmark from human annotators. }
    \label{fig:teaser}
\end{center}%
\bigskip
}]

\begingroup
\renewcommand{\thefootnote}{\fnsymbol{footnote}}
\footnotetext[1]{Equal contribution.}
\endgroup

\begin{abstract}

AI-manipulated video detectors are typically evaluated on clean, curated datasets. In deployment, however, videos have usually passed through repost compression on social-media platforms and sometimes were captured by imperfect hardware. To measure detection under these conditions, we introduce Reality Perception Gap (RPG), a benchmark of 10,000 videos from frontier generators and partially synthetic edits, processed with cross-platform repost compression and paired with real footage that carries camera-side optical artifacts. RPG also includes socially fragile content, such as crime and wartime footage, where fabricated videos can cause severe social harm. Across 18 detectors, none catches more than 15\% of fakes at a 1\% false-positive rate. A positive control and a re-encoding sweep separate two sources of failure: generator and content shift degrades detectors before any compression, and repost compression erases the low-level traces that the artifact-based detectors relies on. Existing detectors therefore struggle when deployment conditions strip away the generation artifacts they rely on. These complementary failures suggest that robust detection requires training under deployment conditions and combining low-level, temporal, and semantic evidence, where passive detection may have reached its limits. The benchmark can be found at \href{https://anonymous-rpg.github.io/RealityPerceptionGap/}{anonymous-rpg.github.io/RealityPerceptionGap/}
\end{abstract}

\section{Introduction}
\label{sec:intro}

Video generation models can now produce photorealistic, temporally coherent clips that once required a professional visual effects studio. They pose a serious threat to the public sphere: synthetic videos that place real people into fabricated scenarios, including non-consensual imagery and politically destabilizing disinformation~\citep{fisher2024moderating}. Such synthetic media already spreads widely on major platforms, whose policies and enforcement treat AI-generated content inconsistently \citep{kira2024nonconsensual, fisher2024moderating}. Because platforms cannot yet reliably distinguish synthetic from authentic content \citep{fisher2024moderating} and takedowns typically arrive only after harm is done \citep{kira2024nonconsensual}, reliable automated detection is needed to curb these harms.

Detection benchmarks have driven progress, but they evaluate detectors under conditions far from how video actually appears in the wild. Most of them either target face manipulations produced by earlier-generation methods \citep{rossler2019faceforensics, dolhansky2020dfdc, yan2024df40}, or cover general-scene generators that predate the current frontier and evaluate predominantly on clean generator outputs, with at most isolated synthetic perturbations \citep{demamba2024, ni2026genvidbench}. 
Wild videos break detector assumptions in two opposite ways. On one hand, repeated social media platform compression strips away the subtle artifacts generative models leave behind that many detector models rely on. On the other hand, videos filmed by imperfect optical hardware introduce distortion that resembles generative artifacts, producing false positives. Existing benchmarks built on clean generator outputs capture neither failure mode, and detectors trained on them already score above 97\%~\citep{demamba2024,ni2026genvidbench}. This leaves little headroom even as these datasets grow to millions of clips, and part of the margin comes from dataset shortcuts, such as frame-rate and codec mismatches between real and generated videos \citep{ben2026training, waverep2024}.

We call this mismatch the reality gap, and argue that closing it is essential for real-world detection. To quantify this gap, we introduce \emph{Reality Perception Gap (RPG)}, a benchmark that prioritizes two underexplored directions: frontier generator coverage and deployment realism. RPG comprises 10,000 videos, fakes from state-of-the-art (SOTA) generators and real footage that includes camera-side optical artifacts, all passed through post-processing that simulates social-media repost compression. Further processing keeps labels from being inferred from file metadata or by human viewers. We withhold the ground-truth labels and score detector predictions through a backend service that returns only aggregate metrics, which prevents label leakage. RPG also includes socially fragile categories such as crime, natural disaster, surveillance, and wartime footage, which carry the social influence when fabricated yet are not represented in existing benchmarks. Using RPG, we evaluate 18 modern detectors: eleven non-VLM detectors spanning frequency and forensic-trace, foundation-feature, and spatio-temporal designs, and seven VLM-based detectors built on vision-language model backbones, five of which explain their decisions in natural language (Section~\ref{sec:discussion}). Under deployment-grounded conditions, every family fails, each in a different way. Our contributions are as follows:
\begin{itemize}
    \item We introduce RPG, a benchmark of 10,000 videos that couples frontier generators and partially synthetic edits with socially fragile content, camera-side optical artifacts, and cross-platform repost chains. Its labels stay on an evaluation server, which keeps the test set free of contamination.
    \item We show that no evaluated detector is deployment-ready: none catches more than 15\% of fakes at the 1\% FPR operating point. The leading detectors, WaveRep~\cite{waverep2024} and D3~\citep{zheng2025d3}, sit at opposite ends of a trade-off between low-level frequency cues and feature-space temporal statistics, and most VLM-based detectors default to ``real'' and miss most fakes.
    \item Through a positive control and a re-encoding sweep, we separate two sources of failure: generator and content shift alone lowers Skyra-RL from 0.74 to 0.63 AUC, and re-encoding drives WaveRep from 0.93 to near 0.5 within 5 re-encodes. No detector handles both shifts, which motivates compression-aware training, semantic reasoning, and proactive watermarking.
\end{itemize}
\section{Related Work}
\label{sec:related}
\begin{figure*}
    \centering
    \includegraphics[width=0.8\linewidth]{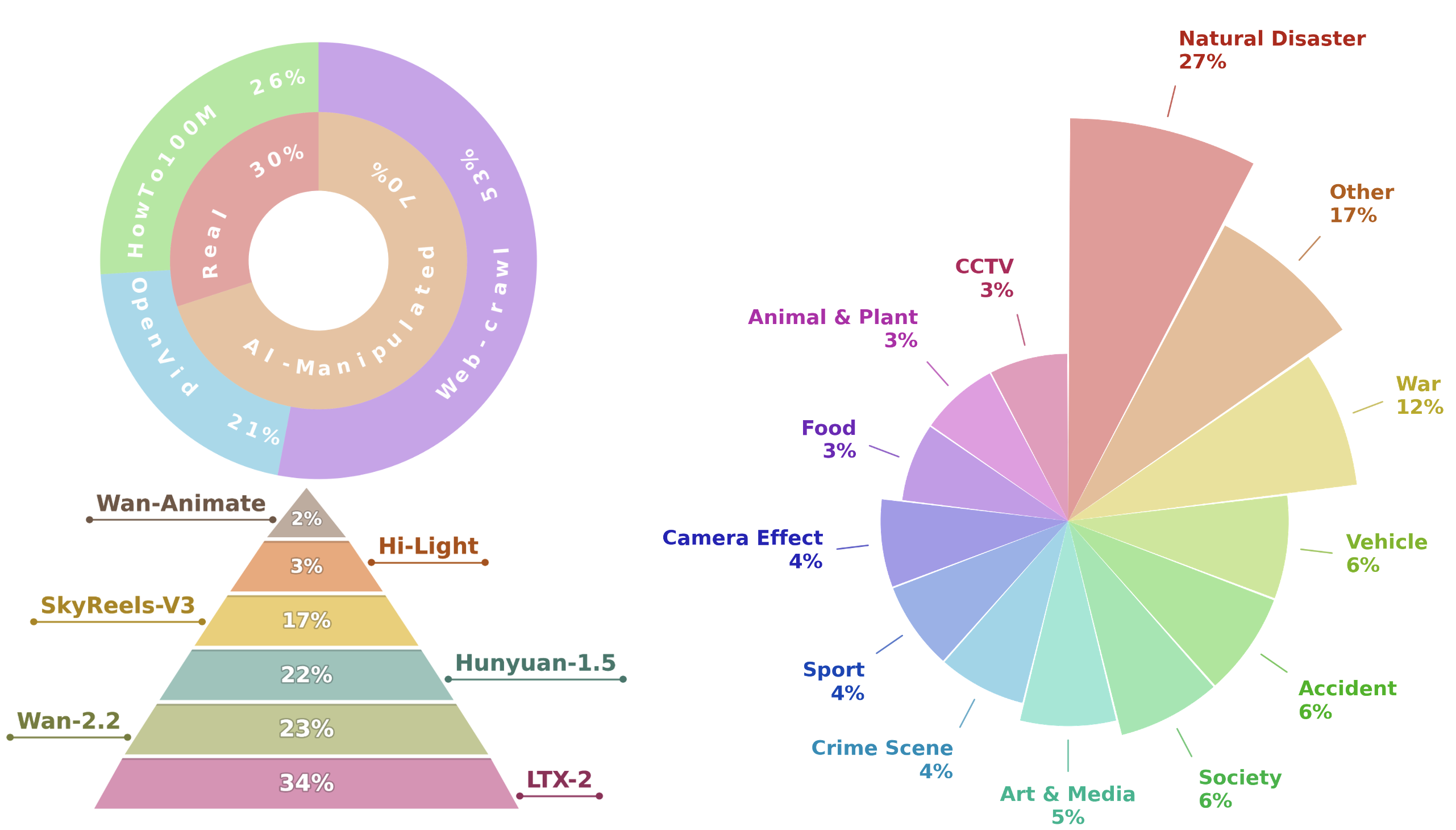}
    \caption{Composition of RPG. Top left: the inner ring shows the label split (30\% real, 70\% AI-manipulated), and the outer ring shows the source dataset of all source videos; the two rings are independent. Bottom left: share of each generator among AI-manipulated videos. Right: content category distribution. Percentages are rounded.}
    \label{fig:data_distribution}
\end{figure*}

\subsection{AI-manipulated Video Detection Methods}
\label{sec:related_methods}

   Detectors differ in the evidence they treat as a trace of generation, ranging from low-level pixel statistics to high-level semantics, and fall into four families along this axis. Frequency and forensic traces modeling. FreqNet~\citep{tan2024freqnet}, FIRE~\citep{chu2025fire}, and WaveRep~\citep{waverep2024} read low-level artifacts left by up-sampling and denoising, through high-frequency representations, reconstruction errors, or wavelet-band augmentation, while VideoFACT~\citep{videofact2024} combines forensic traces with scene context to localize manipulations. These cues transfer across generators but reside in the frequency bands that platform encoders quantize most aggressively. Foundation-feature modeling. UFD~\citep{ojha2023towards}, RINE~\citep{rine2024}, and GenD~\citep{gend2024} train light heads on frozen or minimally tuned pretrained encoders such as CLIP~\citep{radford2021learning}, scoring each frame independently. Spatio-temporal dynamics modeling. AIGVDet~\citep{aigvdet2024} and related detectors~\citep{decof2024,demamba2024} learn motion and cross-frame inconsistency, others use physically or perceptually grounded cues~\citep{restrav2024,nsgvd2024,dfgaze2024}, and the training-free D3~\citep{zheng2025d3}, STALL~\citep{ben2026training}, and SPLIT~\citep{hyun2026split} score second-order feature dynamics, likelihood under a model of real video, and patch-level incoherence, respectively. VLM-based detectors. MM-Det~\citep{song2024mmdet} uses a VLM as a feature extractor with a classification head, and Li et al.~\citep{li2026preserving} build a detector on the Qwen2.5-VL vision encoder that processes videos at their native spatial resolution and temporal duration to preserve high-frequency artifacts. BusterX++~\citep{wen2025busterxpp}, Ivy-xDetector~\citep{jiang2025ivyfake}, Skyra-RL~\citep{li2025skyra}, VideoVeritas~\citep{tan2026videoveritas}, and ARGUS~\citep{zou2026argus} instead pair a real/fake decision with a natural-language explanation. Most of these detectors trade low-level signal access for semantic reasoning; Section~\ref{sec:discussion} measures how this trade holds under compression and generator shift.


\subsection{Existing Benchmarks for AI-manipulated Video Detection}
\label{sec:related_benchmarks}
Existing benchmarks have advanced mainly by scale. Early datasets established protocols on largely pre-2024 generators (GVD~\citep{aigvdet2024}, GVF~\citep{decof2024}, DVF~\citep{song2024mmdet}), GenVideo~\citep{demamba2024} and GenVidBench~\citep{ni2026genvidbench} grew to millions of clips, and AIGVDBench~\citep{ma2026aigvdbench} spans 31 generators and over 440K videos, normalized to H.264 to reduce codec bias~\citep{waverep2024}. AEGIS~\citep{li2025aegis} instead prioritizes difficulty with a hard test set from proprietary models, FakePartsBench~\citep{liu2025fakeparts} targets partially synthetic videos, and GenBuster-200K~\citep{wen2026busterx} adds a Wild track of 150 curated, already-circulating forgeries, the closest prior step toward deployment conditions. Table~\ref{tab:benchmark_comparison} compares these benchmarks with RPG.

Three gaps remain. First, real videos come from general web, action-recognition, or object-tracking corpora (\eg, Kinetics-400, MSR-VTT, OpenVid-1M), and no benchmark samples the high-stakes settings, such as crime, disaster, and conflict, where detection matters most. Second, no benchmark treats deployment conditions as a controlled factor: robustness tests apply isolated perturbations such as JPEG compression or a single H.264 re-encode (\eg, GVD and GenVideo), AIGVDBench and GenBuster-200K normalize every video to one codec setting, and the curated Wild track cannot attribute a failure to a specific transformation; none includes camera-side capture artifacts. Third, beyond codec normalization, no benchmark reports stripping file metadata or neutralizing cues that human viewers could exploit. RPG closes all three gaps: it adds socially fragile and camera-effect footage, varies re-encoding depth as a controlled factor, and strips metadata, normalizes resolution, and masks playback-speed and border cues to keep labels from leaking through file properties or to human annotators (Section~\ref{sec:benchmark}).

\begin{table}[t]
\caption{Comparison with AI-manipulated video detection benchmarks. Part.: partially synthetic edits; Frag.: socially fragile content; Cam.: camera-side optical artifacts; Chain: controlled repeated cross-platform re-encoding; Hidden: test labels withheld. $\circ$: whole-video V2V only. AIGVDBench and GenBuster-200K apply a single codec normalization, and GenVideo tests isolated perturbations (none varies re-encoding as a controlled factor.)}
\label{tab:benchmark_comparison}
\footnotesize
\setlength{\tabcolsep}{4pt}
\begin{tabular}{lccccc}
\toprule
Benchmark & Part. & Frag. & Cam. & Chain & Hidden \\
\midrule
GenVideo~\citep{demamba2024} & $\times$ & $\times$ & $\times$ & $\times$ & $\times$ \\
GenVidBench~\citep{ni2026genvidbench} & $\times$ & $\times$ & $\times$ & $\times$ & $\times$ \\
AEGIS~\citep{li2025aegis} & $\times$ & $\times$ & $\times$ & $\times$ & $\times$ \\
FakePartsBench~\citep{liu2025fakeparts} & \checkmark & $\times$ & $\times$ & $\times$ & $\times$ \\
GenBuster-200K~\citep{wen2026busterx} & $\times$ & $\times$ & $\times$ & $\times$ & $\times$ \\
AIGVDBench~\citep{ma2026aigvdbench} & $\circ$ & $\times$ & $\times$ & $\times$ & $\times$ \\
\midrule
RPG (ours) & \checkmark & \checkmark & \checkmark & \checkmark & \checkmark \\
\bottomrule
\end{tabular}
\end{table}

\section{Benchmark}
\label{sec:benchmark}

Our benchmark addresses these limitations through four design choices that depart from prior benchmarks, each targeting a specific failure mode that current evaluations do not capture:
\begin{itemize}

    \item \textbf{Socially fragile videos.} General-purpose datasets such as HowTo100M~\citep{miech2019howto100m} and OpenVid-HD~\citep{nan2024openvid} mostly contain everyday human activity and instructional content, which is rarely the target of malicious manipulation. We therefore supplement them with self-crawled crime, disaster, surveillance, and wartime footage.
    \item \textbf{Hardware-side optical effects.} We include videos exhibiting auto-focus hunting, rolling shutter, ISO noise, and chromatic aberration, which superficially resemble generative artifacts and test whether detectors can separate optical from generative imperfections.
    \item \textbf{Realistic social-media compression.} Many detectors rely on low-level frequency or pixel cues that platform re-encoding destroys, so every video in RPG passes through simulated cross-platform re-encoding, exposing detectors that depend on uncompressed traces.
    \item \textbf{Partially synthetic content.} RPG includes hybrid videos produced by SOTA editing techniques such as character replacement and relighting, raising the difficulty beyond the real-vs-fully-synthetic regime evaluated by current benchmarks.
\end{itemize}

\paragraph{Video Content.} RPG contains 10,000 video clips, approximately $70\%$ AI-manipulated and $30\%$ real. All source videos, both those kept as real and those used as generation inputs, come from HowTo100M (26\%), OpenVid-HD (21\%), and our YouTube crawl of socially fragile categories (53\%); the content distribution is shown in Figure~\ref{fig:data_distribution}. These categories typically have lower resolution than studio footage, providing natural camouflage against both human and automated evaluators. The camera-effect videos are sourced from publicly available footage exhibiting one of four optical artifacts: \textit{autofocus hunting} (sharpness oscillating as the lens seeks focus), \textit{chromatic aberration} (color fringing at high-contrast edges), \textit{rolling shutter} (skew on fast-moving objects from line-by-line sensor readout), and \textit{high ISO noise} (grainy texture from sensor amplification in low light).

\paragraph{Video Generation} Fakes come from image-to-video (I2V) generation and video-to-video (V2V) editing. For I2V, Skyreels Captioner captions each real video, and Wan-2.2~\citep{wan2025wan}, Hunyuan-1.5~\citep{wu2025hunyuanvideo}, and LTX-2~\citep{hacohen2026ltx} generate 480p and 720p videos from its first frame and caption. For V2V, Wan2.2-Animate~\citep{cheng2025wan} replaces the person in 150 human-centric real videos with one of 50 reference photos from pexels.com, Hi-Light~\citep{liu2026hi} relights 200 real videos under one of ten randomly chosen lighting settings, and Skyreels-V3~\citep{li2026skyreels} extends real videos. We keep relighting at 3\% of the fakes: relit videos carry limited potential for social harm, and a small share prevents a single manipulation type that most detectors fail on from dominating the aggregate results.

\paragraph{Post-processing} Every video then undergoes repost compression: a mixed chain of simulated social-media re-encodes, each applying a platform profile we measured on real uploads to WeChat, WhatsApp, TikTok, RedNote, Instagram, and X (Table~\ref{tab:platform_specs}); this reproduces what audiences see online and erases low-level generation traces. Chains of 1, 5, 10, and 20 re-encodes are assigned in a 4:3:2:1 ratio. Detector AUC changes little beyond 5 re-encodes, up to the 30 we tested (Figure~\ref{fig:positive_control}), so deeper chains add degradation without adding difficulty, and we cap the chains at 20. A green tint, the color cast left when chroma data is corrupted during compression or decoding, is applied to 5\% of the videos. 

We further applied procedures to prevent label leakage or inference by human viewers. First, each real video is cut to a centered window whose length matches a fake generated from the same source video, so clip duration carries no label information. Second, generators output different native frame rates, so normalizing all videos to 24 fps changes their apparent playback speed, a cue human viewers could exploit. Hence, we apply a random 1.3$\times$ speed-up or slow-down to 40\% of all videos, real and fake alike. Third, we center-crop 15\% of the videos to 0.8$\times$ and 15\% to 0.9$\times$: generated videos may show their most visible flaws in the border background, such as distorted secondary objects and figures, and removing the border makes viewers judge the main content. Finally, we strip all container metadata, set every handler name to \texttt{VideoHandler}, and mark the colour description fields as unspecified.

\begin{table}[t]
\centering
\centering
\caption{Per-platform compression profiles used, set from real upload videos
per platform. Codec reports the codec and the encoder profile that gates
the available coding tools; Res.@fps is the highest output resolution and frame rate; CRF is the constant-quality target of our x264/x265 encode
(lower\,$=$\,higher quality).}
\label{tab:platform_specs}
\small
\setlength{\tabcolsep}{3pt}
\begin{tabular}{llcc}
\toprule
Platform        & Codec (profile)  & Res.@fps & CRF \\
\midrule
Instagram       & HEVC (Main)      & 720p@30  & 10 \\
TikTok          & HEVC (Main)      & 1080p@60 & 20 \\
WhatsApp        & H.264 (Baseline) & 720p@60  & 24 \\
RedNote         & H.264 (High)     & 720p@30  & 22 \\
WeChat          & H.264 (High)     & 720p@30  & 22 \\
X (Twitter)     & H.264 (High)     & 720p@30  & 23 \\
\bottomrule
\end{tabular}
\end{table}

    \begin{table*}[t!]
    \centering
    \caption{Quantitative results on RPG, ordered by Macro\,F1 within each group; $\pm$ gives the 95\% bootstrap half-width. A random 70/30 guesser reaches Macro\,F1 0.504, always-fake 0.413, and always-real 0.229, and random scoring gives AUC 0.500 and TPR@1\%FPR 0.010. Human results cover a 22-clip subset with binary labels, so their AUC and TPR@1\%FPR are N/A. VLM-based detectors are scored by log-odds.}
    \label{tab:detector_results_mix}
    \small
    \setlength{\tabcolsep}{5pt}
    \begin{tabular}{l c c c c c}
    \toprule
    Detector & Macro\,F1 & AUC & TPR\textsubscript{fake} & TNR\textsubscript{real} & TPR@1\%FPR \\
    \midrule
    \multicolumn{6}{l}{\textit{Non-VLM detectors}} \\
    D3~\citep{zheng2025d3} & \textbf{0.642}$_{\pm0.010}$ & \textbf{0.693}$_{\pm0.012}$ & 0.739$_{\pm0.010}$ & 0.561$_{\pm0.018}$ & 0.013$_{\pm0.005}$ \\
    SPLIT~\citep{hyun2026split} & 0.563$_{\pm0.011}$ & 0.568$_{\pm0.013}$ & 0.806$_{\pm0.009}$ & 0.316$_{\pm0.017}$ & 0.009$_{\pm0.003}$ \\
    VideoFACT~\citep{videofact2024} & 0.542$_{\pm0.010}$ & 0.573$_{\pm0.012}$ & 0.626$_{\pm0.011}$ & 0.479$_{\pm0.018}$ & 0.011$_{\pm0.005}$ \\
    STALL~\citep{ben2026training} & 0.491$_{\pm0.009}$ & 0.517$_{\pm0.012}$ & 0.513$_{\pm0.011}$ & 0.513$_{\pm0.017}$ & 0.006$_{\pm0.004}$ \\
    FreqNet~\citep{tan2024freqnet} &  0.440$_{\pm0.009}$ & 0.488$_{\pm0.012}$ & 0.349$_{\pm0.011}$ & 0.660$_{\pm0.017}$ & 0.015$_{\pm0.005}$ \\
    FIRE~\citep{chu2025fire} & 0.413$_{\pm0.000}$ & 0.545$_{\pm0.012}$ & \textbf{1.000}$_{\pm0.000}$ & 0.000$_{\pm0.000}$ & 0.012$_{\pm0.005}$ \\
        WaveRep~\citep{waverep2024} & 0.350$_{\pm0.007}$ & 0.638$_{\pm0.011}$ & 0.119$_{\pm0.008}$ & 0.995$_{\pm0.003}$ & \textbf{0.141}$_{\pm0.016}$ \\
    UFD~\citep{ojha2023towards} & 0.266$_{\pm0.005}$ & 0.486$_{\pm0.012}$ & 0.041$_{\pm0.005}$ & 0.965$_{\pm0.006}$ & 0.011$_{\pm0.006}$ \\
    GenD~\citep{gend2024} & 0.245$_{\pm0.003}$ & 0.570$_{\pm0.012}$ & 0.016$_{\pm0.003}$ & 0.990$_{\pm0.004}$ & 0.016$_{\pm0.010}$ \\
    AIGVDet~\citep{aigvdet2024} & 0.236$_{\pm0.003}$ & 0.456$_{\pm0.013}$ & 0.008$_{\pm0.002}$ & 0.975$_{\pm0.006}$ & 0.004$_{\pm0.002}$ \\
    RINE~\citep{rine2024} & 0.233$_{\pm0.002}$ & 0.491$_{\pm0.012}$ & 0.004$_{\pm0.001}$ & \textbf{0.999}$_{\pm0.001}$ & 0.013$_{\pm0.004}$ \\
    \midrule
    \multicolumn{6}{l}{\textit{VLM-based detectors}} \\
    Skyra-RL~\citep{li2025skyra} & 0.478$_{\pm0.008}$ & 0.600$_{\pm0.011}$ & 0.275$_{\pm0.011}$ & 0.975$_{\pm0.005}$ & 0.114$_{\pm0.042}$ \\
    VideoVeritas~\citep{tan2026videoveritas} & 0.435$_{\pm0.009}$ & 0.563$_{\pm0.012}$ & 0.274$_{\pm0.011}$ & 0.820$_{\pm0.014}$ & 0.030$_{\pm0.009}$ \\
    MM-Det~\citep{song2024mmdet} & 0.426$_{\pm0.005}$ & 0.502$_{\pm0.011}$ & 0.956$_{\pm0.005}$ & 0.025$_{\pm0.006}$ &  0.011$_{\pm0.004}$  \\
    Li et al.~\citep{li2026preserving} & 0.402$_{\pm0.008}$ & 0.588$_{\pm0.012}$ & 0.210$_{\pm0.010}$ & 0.883$_{\pm0.012}$ & 0.032$_{\pm0.010}$ \\
    ARGUS~\citep{zou2026argus} &0.357$_{\pm0.008}$ & 0.532$_{\pm0.007}$ & 0.147$_{\pm0.008}$ & 0.917$_{\pm0.010}$ & 0.031$_{\pm0.009}$ \\
    Ivy-xDetector~\citep{jiang2025ivyfake} & 0.356$_{\pm0.008}$ & 0.649$_{\pm0.012}$ & 0.138$_{\pm0.008}$ & 0.946$_{\pm0.008}$ & 0.030$_{\pm0.008}$ \\
    BusterX++~\citep{wen2025busterxpp} & 0.314$_{\pm0.007}$ & 0.510$_{\pm0.013}$ & 0.087$_{\pm0.007}$ & 0.965$_{\pm0.007}$ & 0.040$_{\pm0.012}$ \\
    \midrule
    Human & 0.545 & N/A & 0.445 & 0.669 & N/A \\
    \bottomrule
    \end{tabular}
    \end{table*}
\paragraph{Human Survey}
As a perceptual reference, 212 undergraduate volunteers and 13 AI specialists (graduate students and industry researchers) labeled 22 RPG clips as real or AI-generated. Participation was voluntary, the study collected only per-video labels with no personally identifying or demographic data.


\section{Results and Discussion}

\label{sec:discussion}

We evaluate the 18 detectors in Table~\ref{tab:detector_results_mix}, eleven non-VLM and seven VLM-based, each of which releases code with an official checkpoint or a training-free inference path and runs with its official settings. We score every VLM-based detector by the log-odds $\log p(\text{fake}) - \log p(\text{real})$, read from the answer token for detectors that generate a verbal decision and from the classification head for MM-Det and Li et al, which uses its VLM backbone only as a feature extractor; the log-odds ranks videos identically to the normalized fake probability but does not saturate at high confidence. Binary metrics use each detector's own decision. DeCoF and DeMamba are omitted because neither releases a paper-matched checkpoint or sufficient training details to reproduce one.

RPG contains roughly 70\% fakes, so accuracy rewards trivial predictors: labeling every video as fake already yields about 70\% accuracy, which no evaluated detector exceeds. We therefore report Macro\,F1, AUC, and TPR@1\%FPR, each capturing a different property. Macro\,F1 corrects for class imbalance, AUC measures threshold-free ranking quality, and TPR@1\%FPR measures detection in the strict false-alarm regime that forensic deployment requires, where each false accusation against a real video carries a human cost. The caption of Table~\ref{tab:detector_results_mix} gives the values that trivial predictors reach on each metric, and human performance on a 22-clip subset of RPG (12 AI-manipulated, 10 real) serves as a perceptual reference.

By these metrics, no detector is strong on all three, and the leaders differ across metrics. D3 has the highest Macro\,F1 and AUC, but its real and fake score distributions overlap too much to support a low-FPR operating point, leaving its TPR@1\%FPR at chance. WaveRep catches the most fakes at 1\% FPR, yet at its default threshold it accepts nearly every real video and misses most fakes. The remaining non-VLM detectors, including the training-free SPLIT and STALL, stay below 0.60 AUC, and several fall below chance. Where one of them looks strong on a single metric, as with the near-perfect TNR$_{\text{real}}$ of GenD and RINE or the perfect TPR$_{\text{fake}}$ of FIRE, the score reflects collapse to one class. VLM-based detectors fail in a different way. Most default to ``real'': they accept most real videos but catch only a minority of fakes, while MM-Det collapses toward ``fake''. The strongest members of the family, Ivy-xDetector by AUC and Skyra-RL by Macro\,F1 and TPR@1\%FPR, still trail D3 on AUC and WaveRep at 1\% FPR. Default thresholds therefore span the full range, from highly conservative (WaveRep and most VLM-based detectors) to highly aggressive (FIRE and MM-Det), so deploying any of these detectors requires calibrating its threshold on held-out data.

Human viewers perform at the prevalence floor. On the 22-clip subset, their accuracy (0.547) matches that of labeling every clip as fake (0.545), they flag roughly one in three real clips as AI-generated, and AI specialists do no better than undergraduates (Appendix~\ref{sec:appendix}).

\begin{figure}[t]
    \centering
    \includegraphics[width=1\linewidth]{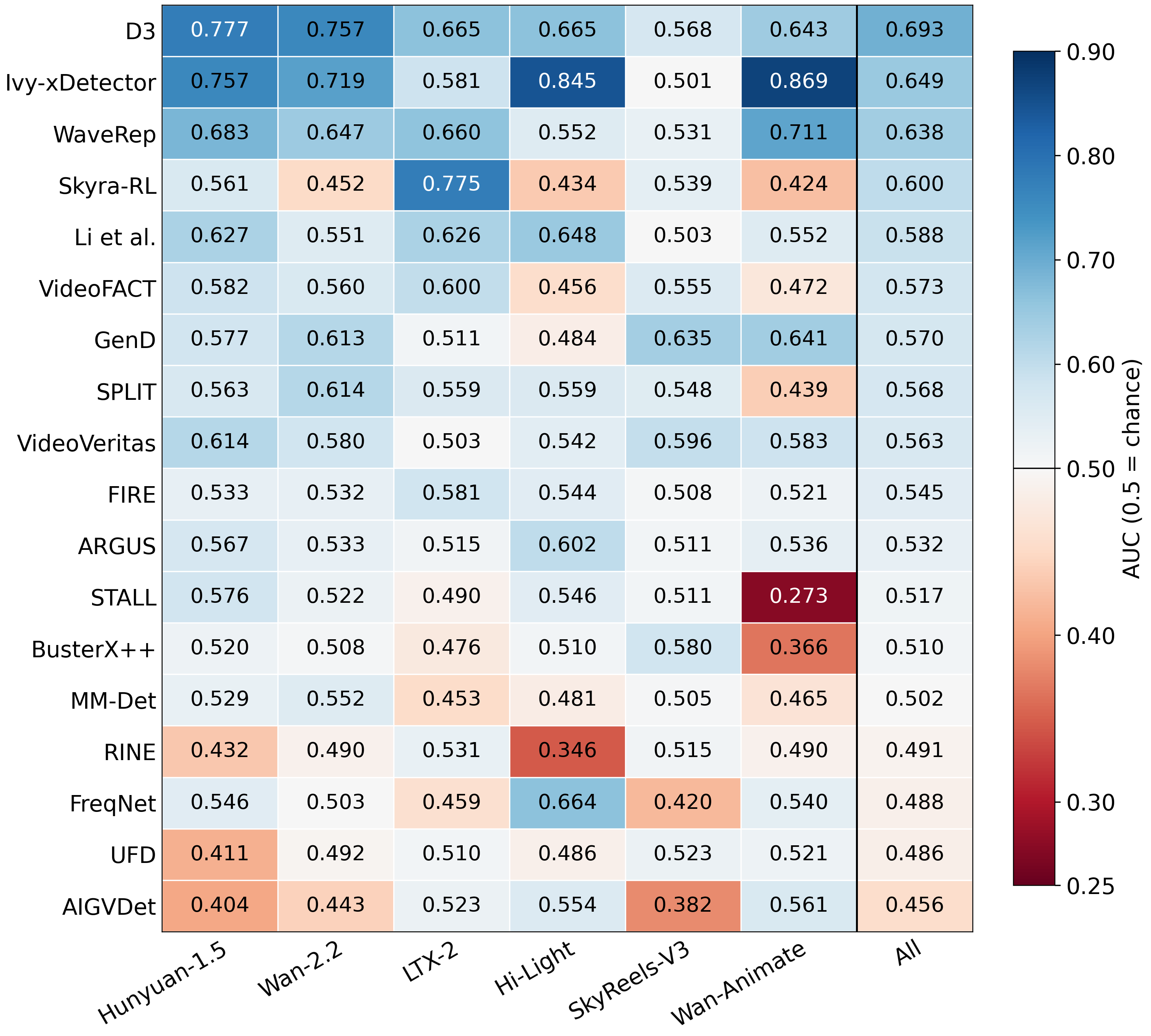}
    \caption{Per-generator AUC of 18 detectors on RPG. Each cell is the AUC of one detector for the fake videos of one generator against all real videos; the All column equals the AUC in Table~\ref{tab:detector_results_mix}. Many detectors stay near chance, and their performance varies widely across generators.}
    \label{fig:heatmap_gen}
\end{figure}

\paragraph{Generator Analysis}
Figure~\ref{fig:heatmap_gen} reports per-generator AUC, and detector performance varies widely across generators. SkyReels-V3 (video extension) is the hardest generator for both WaveRep and D3, and Hi-Light (video relighting) is the second hardest for WaveRep; both produce content where the synthesis residuals WaveRep relies on are weaker and, for SkyReels-V3, the temporal statistics D3 relies on are closer to those of real video. VLM-based detectors are even more generator-dependent: Ivy-xDetector is strong on both partial manipulations, Hi-Light and Wan-Animate, but near 0.5 on SkyReels-V3, and Skyra-RL is strong on LTX-2 but falls below 0.5 on Wan-2.2, Hi-Light, and Wan-Animate. Hunyuan-1.5 is the easiest generator for D3 but pushes UFD, RINE, and AIGVDet below 0.5, indicating that it leaves a distinctive artifact signature that these detectors recognize but classify in the wrong direction.

\begin{figure}[t]
    \centering
    \includegraphics[width=1\linewidth]{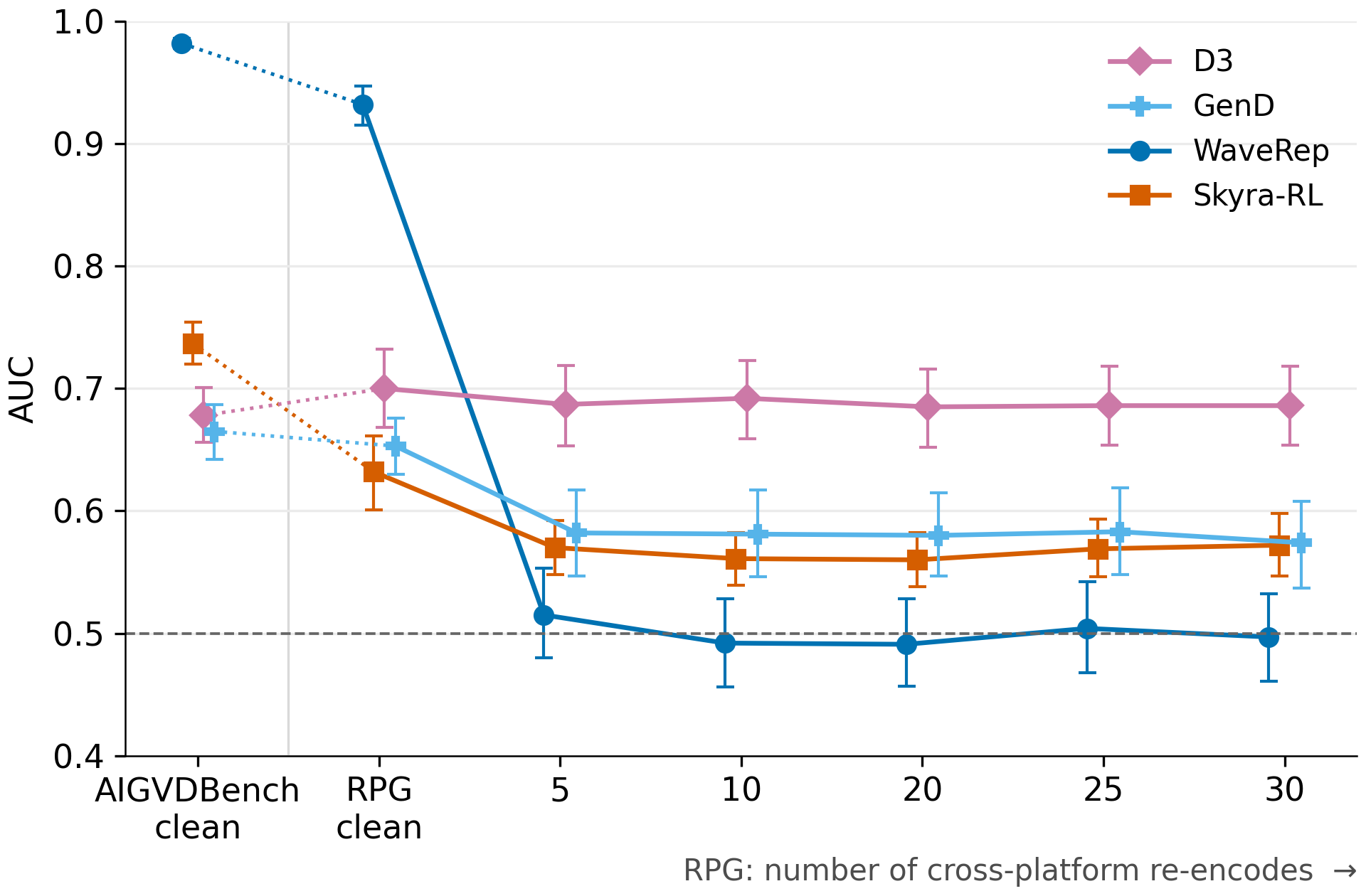}
    \caption{Positive control and re-encoding sweep. Each point is a detector’s AUC on AIGVDBench, or on the same RPG videos  before compression and after 5 to 30 cross-platform re-encodes (right). Each video passes through a mixed chain of platforms.}
    \label{fig:positive_control}
\end{figure}

\paragraph{Positive Control and Re-encoding Sweep}
The positive control (Figure~\ref{fig:positive_control}) uses the strongest detector of each family at the 1\% FPR operating point: WaveRep, GenD, D3, and Skyra-RL. A detector that fails on RPG may fail because RPG is realistic or because the detector never worked; a positive control separates the two. We evaluate the four detectors selected above on AIGVDBench~\citep{ma2026aigvdbench}, on RPG videos before the social media platform compression stage, and on the same RPG videos after up to 30 cross-platform re-encodes. Because the RPG videos are identical across the sweep, every change along the re-encoding axis is caused by re-encoding alone.

\begin{figure*}
    \centering
    \includegraphics[width=0.9\linewidth]{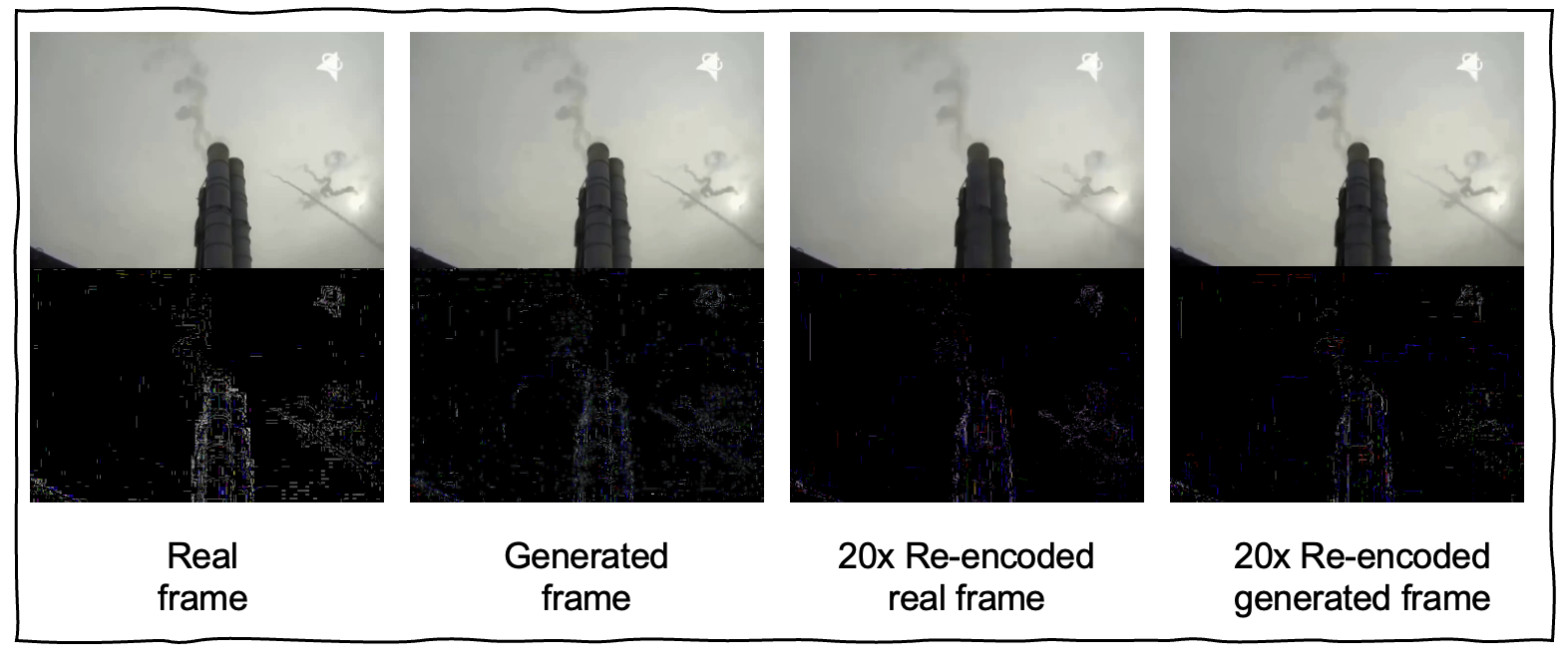}
    \caption{Effect of social-media re-encoding on a generated video. Top: a real frame, the corresponding AI-generated frame, and the corresponding frames after 20 rounds of simulated re-encoding. Bottom: Noise maps to the frames above. Re-encoding removes most background edge responses and keeps the outline of the main object.}
    \label{fig:noise_analysis}
\end{figure*}

Two distinct failures emerge. Generator and content shift alone, with no compression applied, lowers Skyra-RL from 0.74 to 0.63 AUC, while WaveRep stays strong (0.98 to 0.93). Re-encoding reverses this pattern: WaveRep falls to near chance within 5 re-encodes, consistent with platform encoders removing the mid-high frequency residual it reads (Figure~\ref{fig:noise_analysis}), while GenD and Skyra-RL lose a further 0.06 to 0.08. D3 stays near 0.69 under every condition, including AIGVDBench, so it is robust to reposting but never strong. Beyond 5 re-encodes every curve stays flat. No detector is both accurate on clean data and robust to reposting, and RPG measures both failures, which clean benchmarks leave untested.

\paragraph{Detector Mechanisms Analysis.}
Each detector family fails along a different axis of the reality gap. In the frequency and forensic-trace family, WaveRep is the strongest detector at 1\% FPR, but its signal disappears under re-encoding (Figures~\ref{fig:positive_control} and~\ref{fig:noise_analysis}); FreqNet falls below chance, VideoFACT stays weak, and FIRE labels every video as fake. The foundation-feature probes stay below 0.60 AUC, and the strongest, GenD, loses a further 0.08 under re-encoding (Figure~\ref{fig:positive_control}); UFD and RINE fall below chance overall, consistent with a generic ``looks AI'' prior rather than transferable generator fingerprints. In the spatio-temporal family, D3's second-order feature dynamics survive re-encoding nearly unchanged, but its performance varies widely across generators (Figure~\ref{fig:heatmap_gen}) and its default threshold flags nearly half of real videos; AIGVDet, STALL, and SPLIT stay at or below 0.57 AUC. Among VLM-based detectors, Skyra-RL loses much of its margin before any compression is applied (Figure~\ref{fig:positive_control}). No single design covers RPG.

\paragraph{Low-level Noise Analysis}
Figure~\ref{fig:noise_analysis} shows why re-encoding defeats frequency-based detectors. It compares a real frame and the corresponding Wan2.2-generated frame, before and after 20 rounds of simulated social-media re-encoding, together with their noise maps. Before re-encoding, the generated frame carries scattered edge responses over smooth regions such as sky and smoke. After re-encoding, these responses disappear in both frames, and the remaining edges concentrate on the outline of the launcher and on block-aligned codec structures. The two re-encoded frames are therefore nearly indistinguishable at the noise level: repeated re-encoding removes the low-amplitude, high-frequency trace that separates generated from real video and replaces it with codec artifacts common to both, consistent with WaveRep's collapse in Figure~\ref{fig:positive_control}.

\begin{figure*}
    \centering
    \includegraphics[width=0.9\linewidth]{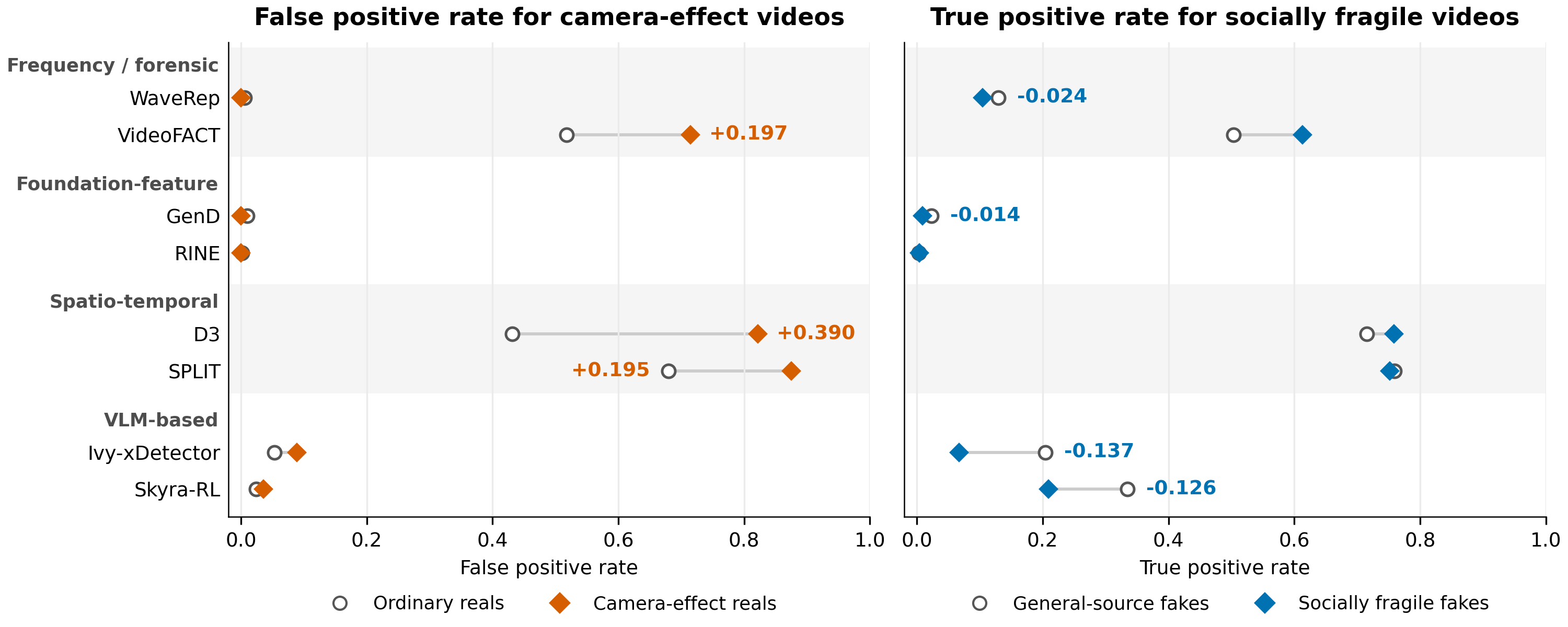}
    \caption{Sensitivity of the top two detectors per family to camera effects and
socially fragile content on RPG, at the released decision thresholds.
\textbf{Left:} false positive rate on real videos with camera effects
and on ordinary real videos.
\textbf{Right:} true positive rate on fake videos with socially fragile content
and on general fake videos.}
    \label{fig:camera_social_plot}
\end{figure*}

\paragraph{Camera effects mislead temporal and forensic cues.}
Figure~\ref{fig:camera_social_plot} compares the top two detectors of each
family on two subsets of RPG at their released decision thresholds. On the
first subset, camera effects can raise false alarms in certain models: real videos with in-camera
effects are flagged as fake far more often than ordinary real videos. The
effect is more observable for the spatio-temporal family, where the false positive
rate rises by 0.390 for D3 and by 0.195 for SPLIT. The forensic detector
VideoFACT shows a similar increase of 0.197. These detectors rely on frame-to-frame dynamics and low-level camera traces, which autofocus hunting, rolling shutter, chromatic aberration, and sensor noise disturb, so authentic footage acquires the signature that the detectors associate with generation. Camera effects are
standard features of smartphone recording, which makes this sensitivity a
direct source of false alarms on everyday user content.

\paragraph{Socially fragile content can hide fakes from VLM-based detectors.}
On the second subset, the failure mode reverses and concentrates in a different
family: VLM-based detectors miss substantially more fakes when the content
depicts CCTV footage, war, crime, accidents or disasters. The true positive
rate drops by 0.137 for Ivy-xDetector, a threefold decrease, and by 0.126 for
Skyra-RL. Both detectors judge authenticity by reasoning about video content,
and we attribute the drop to the low quality, chaotic motion, and plausible
narratives of these scenes, which match how authentic footage of such events looks. This is a red flag for VLM-based
detection: socially fragile events are the scenarios in which fabricated videos
cause the most harm, and they are the scenarios in which these detectors are
least reliable.


\section{Limitations and Future Directions}
\label{sec:limitations}

\textbf{Scope.} RPG reflects the generator and platform landscape of 2026. Because labels are withheld and evaluation runs through a server, the benchmark can be refreshed as generators and platforms evolve without contaminating earlier results. Platform compression is simulated with encoder profiles measured on real uploads, the benchmark is visual only, and the human study covers 22 clips as a perceptual reference.

\textbf{Future Directions.} Our results point to five directions, each grounded in a measured failure. First, detection benchmarks should report performance at a fixed low false-positive rate alongside threshold-free metrics: D3 has the higher AUC, yet WaveRep catches about ten times as many fakes at 1\% FPR (Table~\ref{tab:detector_results_mix}), and default thresholds range from highly conservative to highly aggressive. Second, robustness must be learned under deployment conditions. Even the mid-high frequency cues that WaveRep was designed to preserve do not survive repeated re-encoding (Figures~\ref{fig:positive_control} and~\ref{fig:noise_analysis}). Training on platform re-encoded video and evaluating on an independent corpus directly tests whether this gap closes. Third, the detector families are complementary. WaveRep is the strongest at 1\% FPR, D3 is nearly invariant to re-encoding, and Ivy-xDetector is the strongest detector on both partial manipulations, so combining low-level, temporal, and semantic evidence is a direct path toward a detector that is both accurate and robust. Fourth, semantic reasoning needs calibration and physical grounding. Skyra-RL, the strongest VLM-based detector at 1\% FPR, survives re-encoding far better than WaveRep, losing 0.06 AUC against 0.41. Their confidence, however, is poorly calibrated: most VLM-based detectors answer ``real'' for most videos, yet rank some real videos among their most confident fakes, which keeps the recall of most below 5\% at 1\% FPR. Grounding their decisions in physical and causal plausibility, which generated videos still violate in ways viewers notice, and extending RPG with synchronized audio are natural next steps. Fifth, passive detection has a horizon: a generator whose output matches real video in both low-level statistics and semantics leaves nothing to detect. Proactive authentication, such as a signature embedded in the generator's noise latent that the decoder requires to produce coherent frames, makes removing the signature break the frame. Partial manipulations, where most detectors already degrade because real footage is only relit or locally edited, show where this need is most pressing.
\section{Conclusion}
\label{sec:conclusion}
AI-manipulated video detectors are evaluated under conditions far from deployment, and RPG measures this reality gap with 10,000 videos that combine frontier generators, cross-platform re-encoding, camera-side optical artifacts, socially fragile content, and partially synthetic edits, with labels withheld behind an evaluation server. Across 18 detectors, none catches more than 15\% of fakes at 1\% FPR. Generator shift and re-encoding defeat different detectors, camera effects raise false alarms for temporal and forensic detectors, and socially fragile content can hide fakes from VLM-based detectors, so no current design is robust to all of them. These results call for low false-positive reporting, compression-aware training, detectors that combine low-level, temporal, and semantic evidence, and proactive watermarking where passive detection cannot succeed.

{
    \small
    \bibliographystyle{ieeenat_fullname}
    \bibliography{main}
}

\clearpage
\setcounter{page}{1}
\appendix

\section{Appendix}
\label{sec:appendix}
\subsection{Complete Discussion and Analysis}
\label{sec:appendix_discussion}

This appendix supports Section~\ref{sec:discussion}. It reports the full human survey, the frequency-domain view of re-encoding, additional analysis to the data, and our handling of personal information.

\paragraph{Detailed human survey results.}

\begin{table*}[t]
\small
\centering
\caption{Human survey results for $n=212$ undergraduate respondents. ``Exact'' is the per-respondent rate of selecting exactly the ground-truth fake set. The natural disaster category contains no real clips, so F1 on real clips is undefined.}
\label{tab:human_results}
\setlength{\tabcolsep}{4pt}
\begin{tabular}{lcccccccc}
\toprule
Category & \makecell{Fake/\\Real} & Acc & \makecell{Precision\\(fake)} & \makecell{Recall\\(fake)} & \makecell{F1\\(fake)} & \makecell{F1\\(real)} & \makecell{Macro\\(F1)} & Exact\% \\
\midrule
Relit video & 1/2 & 0.587 & 0.413 & 0.571 & 0.479 & 0.657 & 0.568 & 25.47\% \\
Camera optical effect & 1/3 & 0.565 & 0.234 & 0.325 & 0.272 & 0.690 & 0.481 & 12.74\% \\
Socially fragile content & 2/2 & 0.592 & 0.603 & 0.538 & 0.569 & 0.613 & 0.591 &  9.43\% \\
Compression & 2/2 & 0.653 & 0.692 & 0.552 & 0.614 & 0.685 & 0.650 & 18.87\% \\
Natural disaster & 3/0 & 0.502 & 1.000 & 0.502 & 0.668 & n/a   & n/a   & 13.21\% \\
Extended video & 3/1 & 0.396 & 0.750 & 0.292 & 0.421 & 0.369 & 0.395 &  1.89\% \\
\midrule
\textbf{Overall} & \textbf{12/10} & \textbf{0.550} & \textbf{0.618} & \textbf{0.455} & \textbf{0.524} & \textbf{0.572} & \textbf{0.548} & --- \\
\bottomrule
\end{tabular}
\end{table*}

\begin{table*}[t]
\small
\centering
\caption{Human survey results for $n=13$ AI-specialist respondents. ``Exact'' is the per-respondent rate of selecting exactly the ground-truth fake set. The natural disaster category contains no real clips, so F1 on real clips is undefined.}
\label{tab:graduate_results}
\setlength{\tabcolsep}{4pt}
\begin{tabular}{lcccccccc}
\toprule
Category & \makecell{Fake/\\Real} & Acc & \makecell{Precision\\(fake)} & \makecell{Recall\\(fake)} & \makecell{F1\\(fake)} & \makecell{F1\\(real)} & \makecell{Macro\\(F1)} & Exact\% \\
\midrule
Relit video & 1/2 & 0.667 & 0.500 & 0.538 & 0.519 & 0.745 & 0.632 & 38.46\% \\
Camera optical effect & 1/3 & 0.635 & 0.200 & 0.154 & 0.174 & 0.765 & 0.470 & 15.38\% \\
Socially fragile content & 2/2 & 0.673 & 0.737 & 0.538 & 0.622 & 0.712 & 0.667 &  7.69\% \\
Compression & 2/2 & 0.538 & 0.556 & 0.385 & 0.455 & 0.600 & 0.527 &  0.00\% \\
Natural disaster & 3/0 & 0.308 & 1.000 & 0.308 & 0.471 & n/a   & n/a   &  0.00\% \\
Extended video & 3/1 & 0.288 & 0.583 & 0.179 & 0.275 & 0.302 & 0.288 &  0.00\% \\
\midrule
\textbf{Overall} & \textbf{12/10} & \textbf{0.521} & \textbf{0.612} & \textbf{0.333} & \textbf{0.432} & \textbf{0.586} & \textbf{0.509} & --- \\
\bottomrule
\end{tabular}
\end{table*}

Tables~\ref{tab:human_results} and~\ref{tab:graduate_results} report the human survey for 212 undergraduate and 13 AI-specialist respondents on the same 22 clips (12 AI-manipulated, 10 real). Overall accuracy is similar for the two groups (0.550 and 0.521), and specialists perform slightly worse on fake recall (0.333 vs.\ 0.455) and Macro F1 (0.509 vs.\ 0.548), so on this clip set domain expertise did not improve detection. No specialist identified the exact set of fake clips in the compression, natural disaster, or extended video categories. Extended video is the only category where both groups fall below chance (accuracy 0.396 and 0.288), and camera-effect and extended videos have the lowest fake recall for both groups (0.325 and 0.292 for undergraduates; 0.154 and 0.179 for specialists). Extended video is also among the hardest cases for the two strongest detectors, WaveRep and D3 (Figure~\ref{fig:ablation_sk3}). Together, these results show that the selected clips are challenging for both lay and expert viewers. Because each category contains only one to four clips and the specialist group is small, we treat category-level results as indicative rather than conclusive.

\begin{figure}[h!]
    \centering
    \includegraphics[width=1\linewidth]{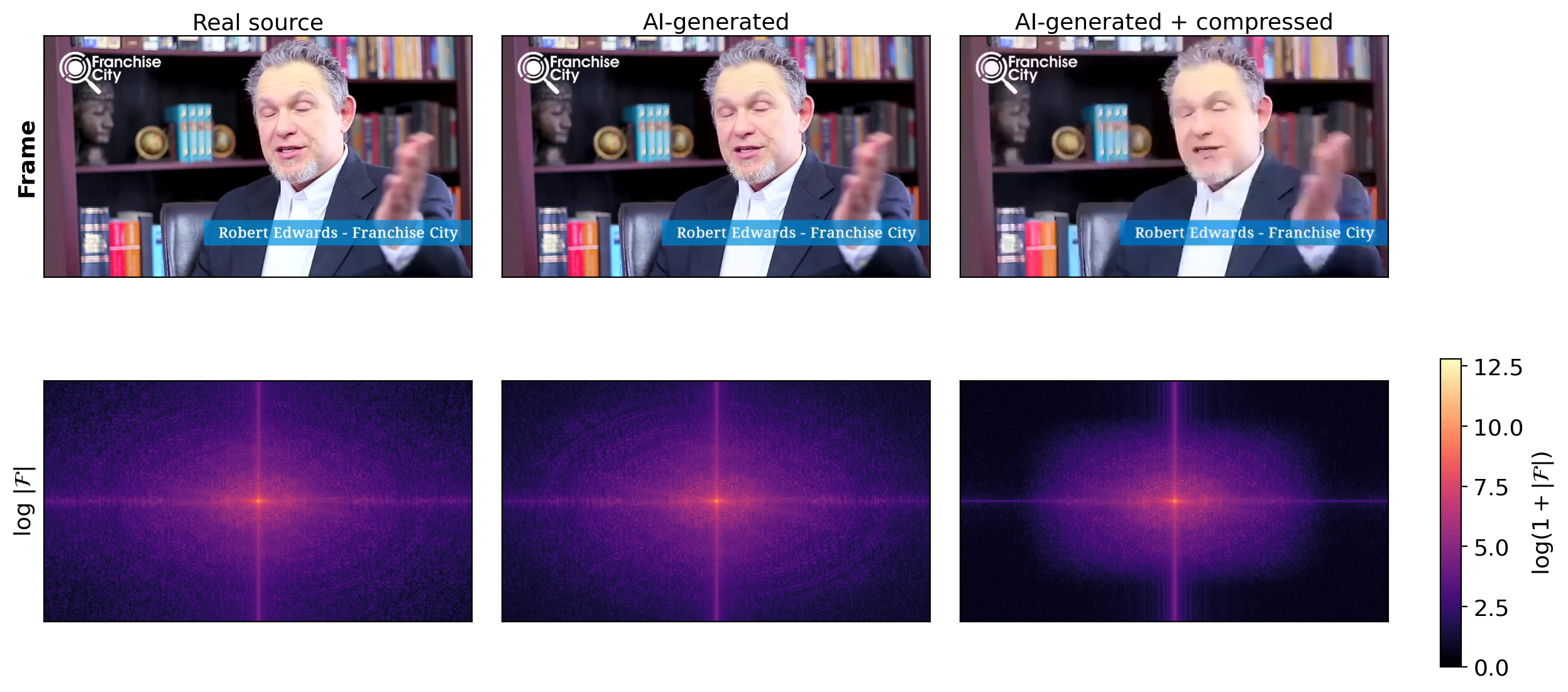}
    \caption{Frequency-domain effect of platform compression on AI-generated video. Top row: a real frame, the corresponding Wan2.2 frame, and the same generated frame after ten rounds of social media compression. Bottom row: log-magnitude 2-D FFT of each frame's luminance channel. The high-frequency content visible at the corners of the AI-generated spectrum, the band that frequency-domain detectors are trained to read, is largely absent after re-encoding, leaving a spectrum that more closely resembles a real frame's than the generator's own uncompressed output.}
    \label{fig:frequency}
\end{figure}

\paragraph{Frequency content and compression.}
Figure~\ref{fig:frequency} gives the spectral view of the mechanism in Figure~\ref{fig:noise_analysis}. After ten rounds of social media compression, the high-frequency band of a Wan2.2-generated frame largely disappears, leaving a spectrum closer to a real frame's than to the generator's own output. FreqNet, which applies frequency-domain convolutions to high-frequency representations, loses its input entirely and stays at chance on RPG (Table~\ref{tab:detector_results_mix}). WaveRep is trained toward mid-high frequency diagonal artifacts, which survive a single compression pass better than horizontal and vertical components but still fade as re-encodes accumulate, so its AUC reaches chance within 5 re-encodes (Figure~\ref{fig:positive_control}). Frequency cues alone are therefore unreliable under repeated redistribution, which motivates training detectors under deployment conditions (Section~\ref{sec:limitations}).

\subsection{Generator Analysis}
Each analysis compares a detector's AUC on real videos against the fakes of one generator with its AUC against the fakes of all other generators, both computed on the same real videos.

\begin{figure}[t]
    \centering
    \includegraphics[width=1\linewidth]{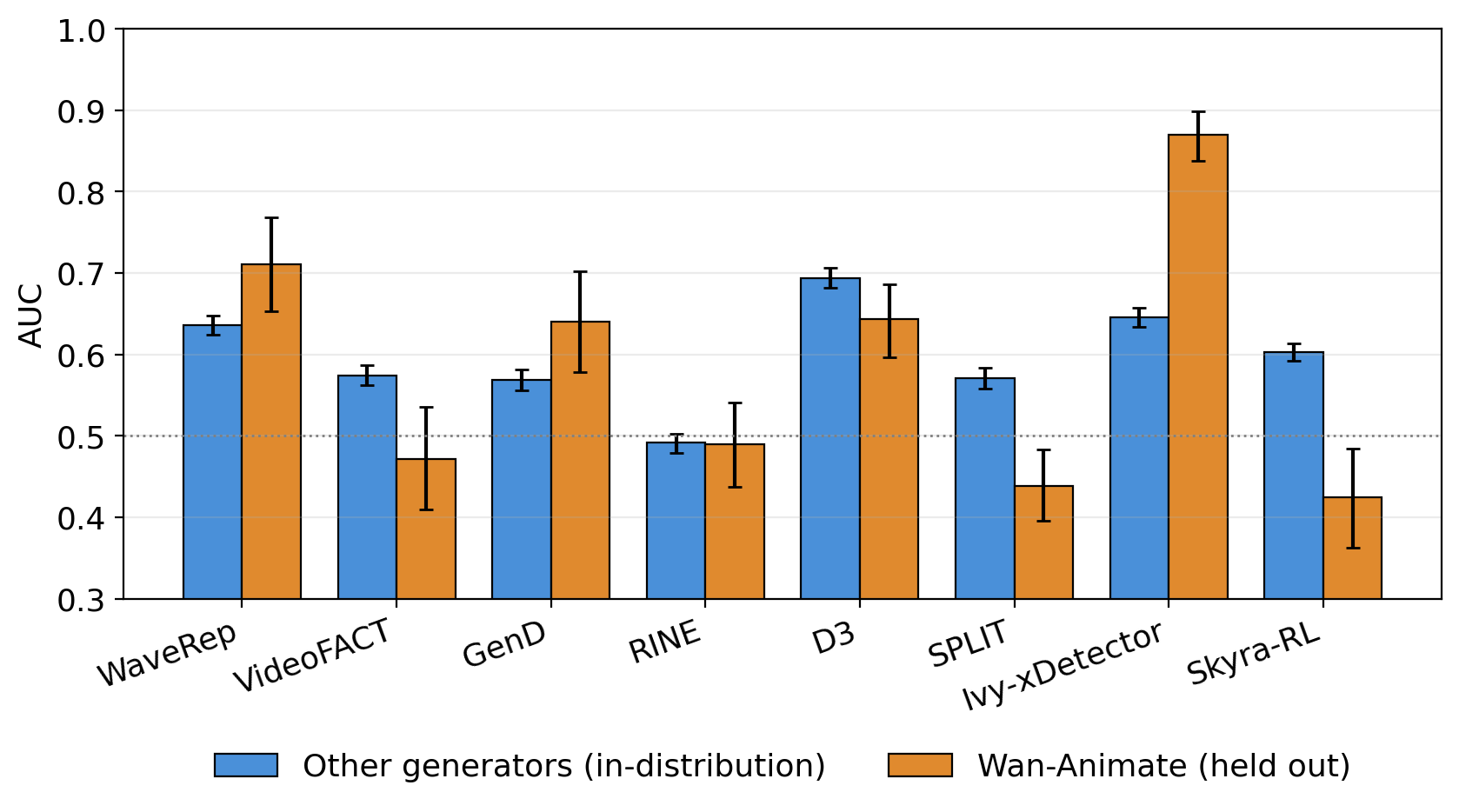}
    \caption{Wan-Animate character replacement analysis. Per-detector AUC on real videos against Wan-Animate fakes, compared with AUC against the fakes of all other generators.}
    \label{fig:appendix_ood_wan_animate}
\end{figure}

\paragraph{Character replacement analysis.}
Figure~\ref{fig:appendix_ood_wan_animate} isolates Wan-Animate, our
character-replacement generator, and splits the detectors into two groups.
Ivy-xDetector, GenD, and WaveRep detect Wan-Animate fakes more reliably than
fakes from the other generators, with Ivy-xDetector showing the largest gain.
D3 drops slightly, while VideoFACT, SPLIT, and Skyra-RL fall to or below
chance; RINE stays near chance in both cases. The wider error bars on the
held-out bars reflect the smaller Wan-Animate subset. This split is consistent
with what Wan-Animate changes and what it preserves: it renders a synthetic
character into real footage, so the background, camera motion, and scene
dynamics keep real-world statistics, while the synthesis artifacts concentrate
on the replaced character. Detectors that respond to the appearance of the
synthesized region benefit from this concentrated signal, whereas detectors
that rely on scene-level or temporal cues see a video that is largely real.
Generator difficulty is therefore not a single axis: what a generator changes
and what it preserves determines which detectors it evades.

\begin{figure}[t]
    \centering
    \includegraphics[width=1\linewidth]{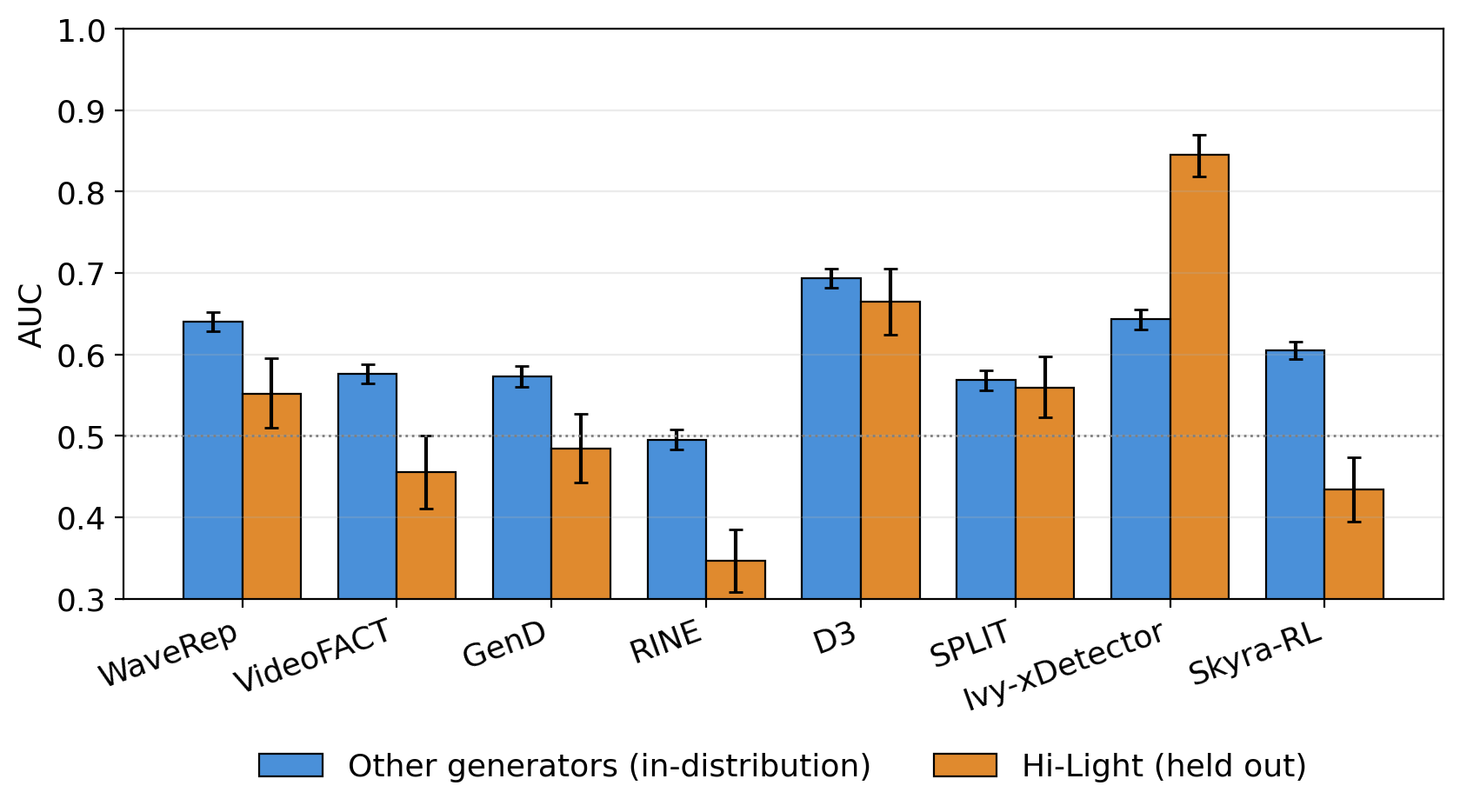}
    \caption{Hi-Light only relights existing footage rather than synthesizing new content, so most detectors degrade, D3 and SPLIT are nearly unchanged, and Ivy-xDetector improves sharply.}
    \label{fig:ablation_relight}
\end{figure}

\paragraph{Video relighting analysis.}
The Hi-Light analysis in Figure~\ref{fig:ablation_relight} gives the most
pessimistic picture of the three probes. Most detectors degrade on relit
videos: WaveRep, VideoFACT, and GenD lose much of their margin, and RINE and
Skyra-RL drop well below chance. D3 and SPLIT are nearly unchanged. Hi-Light
keeps the real source footage and modifies only its lighting and shading, so
little generated content remains to fingerprint, and detectors that rely on
spatial, frequency, or temporal traces of synthesis lose their primary signal.
Ivy-xDetector is the clear exception: it detects relit videos far more
reliably than fakes from the other generators and reaches the highest
per-generator AUC in the analysis, which indicates that it registers
lighting-induced inconsistencies that the other detectors miss. Manipulations
that perturb a real video instead of synthesizing a new one are therefore
especially hard for current passive detectors.

\begin{figure}[t]
    \centering
    \includegraphics[width=1\linewidth]{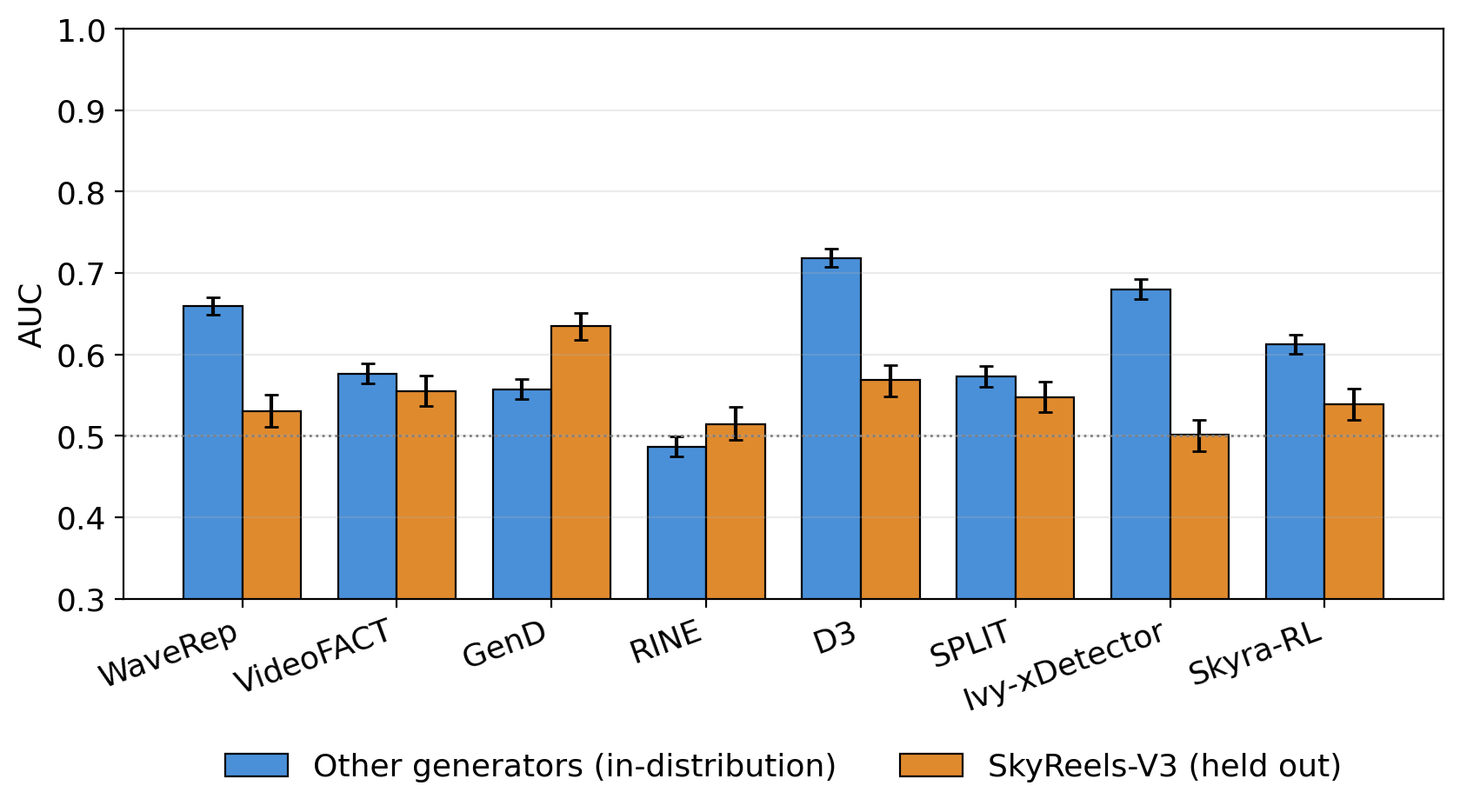}
    \caption{Extended video analysis. The detector ranking partially inverts on this generator: the four strongest detectors overall drop sharply, while GenD and RINE improve.}
    \label{fig:ablation_sk3}
\end{figure}

\paragraph{Video extension Analysis.}
The SkyReels-V3 analysis in Figure~\ref{fig:ablation_sk3} shows the largest consistent effect in the appendix and a partial reversal of the detector
ranking. The four strongest detectors by overall AUC (D3, Ivy-xDetector,
WaveRep, and Skyra-RL) all drop sharply on SkyReels-V3 videos, and
Ivy-xDetector falls to chance. VideoFACT and SPLIT decline slightly, while
GenD and RINE move in the opposite direction, with GenD showing a clear gain.
The strongest detectors on the rest of the benchmark thus lose most of their
advantage, while several weaker ones improve. One explanation is that
SkyReels-V3's long-horizon outputs accumulate frame-level inconsistencies,
such as texture and semantic drift, that the weaker detectors can read, while
the generator's long-context conditioning smooths the cues that drive the
strongest detectors. This is the strongest case in our results for
complementary detector design: no single design dominates across the three manipulation types analyzed here.
\subsection{Handling of Personal Information}
RPG contains real videos crawled from publicly available web sources covering crime, disaster, and conflict events. All videos are distributed without audio tracks. The following sensitive categories may be incidentally present in the visual content: geography (videos depict identifiable real-world locations), language (incidental on-screen text, predominantly English with some other languages), age and gender (individuals of varying ages and genders may appear incidentally as subjects of news or public-incident footage), and culture (videos reflect the cultural contexts of their source events and outlets). The benchmark does not intentionally collect or label health or medical data, socio-economic status, political or religious beliefs, or experience or seniority information. No biometric identifiers such as face embeddings or voiceprints are extracted or distributed, and the absence of audio tracks removes any spoken-content sensitivity.

All real videos are sourced from publicly accessible web content originally published by news outlets, social platforms, or public archives; no private or restricted-access material is included. To protect individuals whose identification could carry heightened risk, we apply mosaic masking to identifiable soldiers and other military or security personnel visible in conflict footage, since identification of such individuals can have direct consequences for their safety. We do not add identifying metadata beyond what is already public, and we do not perform face re-identification or any analysis intended to identify specific individuals. Synthetic videos contain no real personal information by construction, though generator outputs may include synthesized faces resembling real demographic groups.

Residual risk: because real videos depict real public events, individuals visible in the footage may be identifiable to viewers familiar with those events. This is inherent to the source material and matches the level of identifiability already present in the original publications. Mosaic masking is applied where identification carries heightened personal risk but is not applied universally to all visible faces, since blanket masking would alter the visual statistics the benchmark is designed to evaluate.

\subsection{Use of Large Language Models (LLMs)}
LLMs were used to help visualize results, correct grammar, and understand technical concepts.

\end{document}